\documentclass[a4paper, 12pt, one column]{article}

\usepackage[english]{babel}
\usepackage[utf8x]{inputenc}
\usepackage[T1]{fontenc}

\usepackage[top=1.3cm, bottom=2.0cm, outer=2.5cm, inner=2.5cm, heightrounded,
marginparwidth=1.5cm, marginparsep=0.4cm, margin=2.5cm]{geometry}

\usepackage{graphicx} 
\usepackage[colorlinks=False]{hyperref} 
\usepackage{amsmath}  
\usepackage{amsfonts} %
\usepackage{amssymb}  %

\newtheorem{definition}{Definition}

\usepackage[authoryear]{natbib}
\setcitestyle{authoryear,open={(},close={)}}

\title{Dendrogram distance: an evaluation metric for generative networks using hierarchical clustering}

\author{Gustavo Sutter P. Carvalho, Moacir A. Ponti}

\begin{document}
\maketitle

\begin{abstract}
We present a novel metric for generative modeling evaluation, focusing primarily on generative networks. The method uses dendrograms to represent real and fake data, allowing for the divergence between training and generated samples to be computed.
This metric focus on mode collapse, targeting generators that are not able to capture all modes in
the training set. To evaluate the proposed method it is introduced a validation scheme based on
sampling from real datasets, therefore the metric is evaluated in a controlled environment and
proves to be competitive with other state-of-the-art approaches.
\end{abstract}

\section{Introduction}
\label{intro}

\begin{figure}[h]
\caption{Diagram that illustrates the different components of GAN. The generator network $G$ transforms a random input $z$ into samples that should be realistic, while the discriminator network $D$ tells apart which samples came from the training data.}
\label{fig:gan}
\centering
\includegraphics[width=0.5\textwidth]{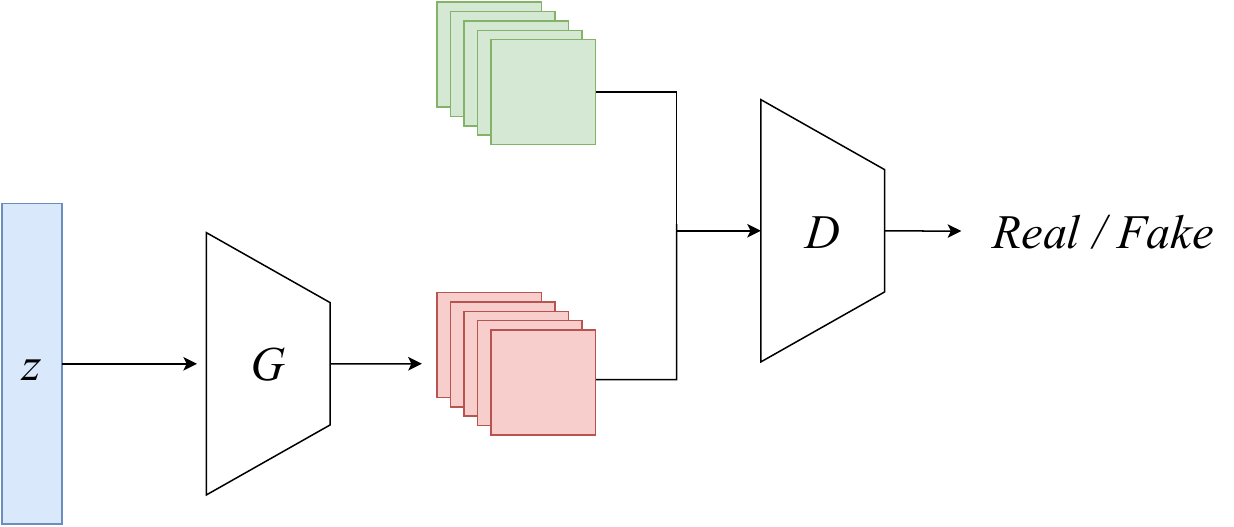}
\end{figure}

Generative modeling is a task that aims to estimate the generation process of a given source dataset. Models obtained as a result of this approach can be used to sample novel data points that follow the distribution of the source training set, allowing for different applications in machine learning. Performing generative modeling using neural networks has become very popular mainly because of the success of Generative Adversarial Networks (GANs)~\citep{goodfellow2014} and later with Diffusion models~\citep{luo2022understanding}. The GAN framework relies on two different networks, a generator and a discriminator, that compete against their selves to perform the generative task, as shown in Figure \ref{fig:gan}.

More specific, a GAN is composed by a generator network $G$ and a discriminator network $D$. The goal of the generator is to produce realistic samples, while the discriminator acts as a binary classifier telling apart samples from the training set and created by the generator. That is to say that $G$ receives a random vector $z$ as input and tries to approximate the real data distribution $p_{data}$ using a distribution $p_{model}$. The whole system is trained together, optimizing the following mini-max objective:
\begin{equation}
    \min_G\max_D L(G, D) = \mathbb{E}_{x \sim p_{data}}[\log D(x))] + \mathbb{E}_{\tilde x \sim p_{model}}[\log (1 - D(G(\tilde x))],
\end{equation}
where $x \in X$ is sampled from a training set $X$ containing $n$ examples, while $\tilde x$ comes from the generator.

The field is prolific, which can be seen by an exponential number of published papers in the last  years~\citep{faria2020generative}. Its approach, which uses two neural networks trained against each other, demonstrated to produce great results even in hard high dimensional settings. Because of that, many other methods built on top of the traditional adversarial setup were developed improving its results and contribution to its use.

Even though GANs have achieved success in different applications, such as image super-resolution~\citep{ledig2017}, style transfer~\citep{zhu2017}, image synthesis~\citep{wang2018}, sketch synthesis~\citep{sampaio2022scene} and tracking~\citep{comes2020multi}, the difficulty of training and evaluating a generative model remains its main limitations~\citep{ponti2021training}. With respect to training, instability of the generator results and the high sensitivity to network architecture and hyper-parameters are among the relevant issues.  Even when the model converges, there are notably two possible undesired effects: low sample quality and mode collapse~\citep{adiga2018tradeoff}.  Keeping the problem of generating images in mind, low sample quality happens when the generated images are blurry, noisy or with undesirable visual artifacts. Mode collapse, on the other hand, has to do with how diverse the generated data is. When the data produced by the generative model fails to cover the modes of the training data it said that it suffers from mode collapse.

The substantial progress of applications and generative approaches are not equally followed by efforts on evaluating them~\citep{borji2019pros, ribeiro2020sketchformer}. However, even those effects are not easy to be measured. In contrast with supervised learning methods, such as classification, developing metrics for generative modeling is not as straightforward because of the ill-posed nature of the task. More specifically, as the task is to produce samples that follow the training data distribution, there is no direct and trivial way to determine how good a sampled instance is~\citep{parimala2019quality}~\citep{shmelkov2018good}. This issue is amplified when dealing with high dimensional data as images.

Several metrics have been proposed on the last years, notably the Inception Score~\citep{salimans2016improved}~\citep{gurumurthy2017deligan} and the Frechet Inception Distance~\citep{heusel2017gans}, which are widely used and currently the de-facto evaluation references. Both metrics make use of a pre-trained convolutional network to extract a latent representation in order to deal with high-dimensional image data. Some problems regarding this metrics are well known in the literature~\citep{barratt2018note}~\citep{xu2018empirical}~\citep{borji2019pros} as we present evidence that both can fail to detected mode collapse on different datasets.

We propose the Dendrogram Distance (DD), a novel evaluation metric for generative models which relies on hierarchical clustering to evaluate the generated data. Using results from clustering theory, we demonstrate the equivalence between dendrograms and ultrametric spaces to develop a metric that has theoretical foundations~\citep{carlsson2010characterization} and that is sensitive to variations in the generated data, in particular to detect generated modes. Our method captures the ordered hierarchical clustering via dendrograms and is guaranteed to detect differences when they exist. This guarantee is not present in previous evaluation methods, therefore it represents a relevant contribution.

Our experiments demonstrate that Dendrogram Distance is capable of detecting different number of modes on generated data. In particular, it was shown to be sensible to nuances of the space for which the competing methods fail to capture, such as when the modes' means are not equally spaced. Therefore, this makes DD a superior choice to 2D benchmark, but also useful to evaluate directly the generated data, such as image pixels or image embeddings, as well as capturing convergence of models. 

\section{Related work}
\label{sec:related}

Generative deep learning has became very popular since the introduction of Generative Adversarial Networks (GANs)~\citep{goodfellow2014}. This framework, composed of two neural networks that compete against each other, rapidly became the state-of-the-art in various tasks over the past years, being applied to a variety of tasks. However, the evaluation of generative models is a challenging task since there is no explicit label or category to compare with, such as in supervised learning. Grasping the quality and diversity of generated data is often performed visually, in particular when it comes to early studies on deep generative methods~\citep{radford2016}.

Evaluating generative methods is a challenge and, as of yet, there is no consensus on the best way to to it~\citep{borji2019pros}. Notably the most widely used evaluation method for image generation is the Inception Score (IS) \citep{salimans2016improved}, which uses a pre-trained Inception-V3 \citep{szegedy2016rethinking} to verify if the samples produced by the generative model present high quality and diversity. To this end the IS uses the class distribution $p(y|x)$ produced by the network when passing through it the generated samples. The main point is that high quality samples must imply that images should be easily identifiable, therefore $p(y|x)$ must have low entropy, and given a set a images its marginal distribution $p(y)$ must have high entropy, meaning that there must diversity. Some variations on the IS were proposed but are not as widely used, for example: the Mode Score~\citep{che2016mode}, AM Score~\citep{zhou2017activation} and Modified Inception Score~\citep{gurumurthy2017deligan}.

Another metric popularly used for evaluating generative modeling on images is the Fréchet Inception Distance (FID)~\citep{heusel2017gans}. Simliarly to the the Inception Score the FID also makes use of a pre-trained Inception-V3 as a feature extractor, but it used the activations from the last convolutional layer to represent the data points. It assumes that these representations, for the training and generated sets, follow a multivariate normal and compute the Fréchet Distance between the two. That is done by simple estimating their mean vector and covariance matrices. Thus, FID operates on the feature spaces and therefore can be applied to non-image data.

One important drawback of IS and FID scores is that they make more sense in large scale natural scene datasets, while for other tasks one often has to design specific evaluation protocols~\citep{yang2020evaluation}. Both the IS and FID can be categorized as extrinsic evaluation metrics, given that they measure the performance of the generator based on a downstream task, namely how good are the representations produced by a network pre-trained with data from the same domain. Other approaches for image data rely on image quality metrics such as MS-SSIM~\citep{odena2017conditional}, that uses structural-similarity for image pixels. On the other hand there is a set of metrics that do not rely on external models or data, which are referred intrinsic evaluation metrics~\citep{theis2016note}. 

A standard approach to perform intrinsic evaluation of a generative model is to estimate its density, via Kernel Density Estimation, and compute classical metrics such as the Kullback–Leibler or Jensen–Shannon divergences. However, this method requires too much data when applied to high-dimensional spaces and relies on the Euclidean distance between data points due the kernel probability, which can be harmful when dealing with images~\citep{theis2016note}.  

All the aforementioned metrics result in a number that allow to compare different generative results, however, specific characteristics such as mode collapse and sample quality are not explicitly addressed. Therefore, 2D benchmarks became a common approach, i.e. the use of synthetic datasets that allow for rapid investigation of the results. The 2D Ring and the 2D Grid datasets can be used to visualize such aspects \citep{srivastava2017veegan} and makes it particularly easier verify how well a generative model does in terms of mode collapse. It provides useful feedback when iterating over ideas and a controlled and explainable benchmark. Although it may seen to be an easy task because of its low dimensionality, 2D datasets represent a great challenge to generative models such as GANs. Those are a way to clearly visualize how the model covers the target modes.

\section{Method}
\label{sec:method}

We propose an evaluation method based on the divergence between dendrograms. Our approach follows from the natural assumption that if the generated data is similar to the real data, the clustering of their samples must also be similar. The hypothesis is that a dendrogram allows the metric to better capture the distributions than the first and second moments.

\subsection{Hierarchical Clustering and Dendrograms}

\begin{definition}
If $X$ is a finite set, $\mathcal{C}(X)$ is the collection of all non-empty subsets of $X$
\end{definition}
\begin{definition}
If $X$ is a finite set, $\mathcal{P}(X)$ is the collection of all partitions of $X$
\end{definition}

A dendrogram over a dataset $X$ is a representation of the hierarchical clustering of the data, often represented as a tree. Formally, dendrograms are described as a pair $(X, \theta)$, where $X$ is the dataset and $\theta: [0,\infty] \rightarrow \mathcal{P}(X)$ is a function that maps every distance to the set of clusters at that point, that holds the following properties~\citep{costaemploying}:
\begin{enumerate}
    \item $\theta(0) = \left\lbrace \left\lbrace x_0 \right\rbrace, \left\lbrace x_1 \right\rbrace, \ldots \left\lbrace x_n \right\rbrace \right\rbrace$: the initial level of the dendrogram is the most fine grained representation as possible;
    \item There is tome $t_0$ such that $\theta(t)$ is a unique block for every $t \geq t_0$, so that for every value $t > t_0$ dendrograms will be the same and contain all elements, i.e. $\theta(t) = \left\lbrace \left\lbrace x_0, x_1, \ldots x_n \right\rbrace \right\rbrace$;
    \item If $r \leq s$ then $\theta(s)$ is a more fine grained dendrogram than $\theta(r)$, which ensures a sense of scale for $\theta(.)$ and suggests nested partitions;
    \item For all $r$ there is an $\varepsilon > 0$ such that $\theta(r) = \theta(t)$ for $t \in [r, r+\varepsilon]$.
\end{enumerate}

\begin{figure}[h]
\caption{A dendrogram built over the set $X = \{x_1, x_2, x_3, x_4\}$. In this example the agglomerative distances are $d = [2, 4, 6]$.}
\label{fig:dendro}
\centering
\includegraphics[width=0.35\textwidth]{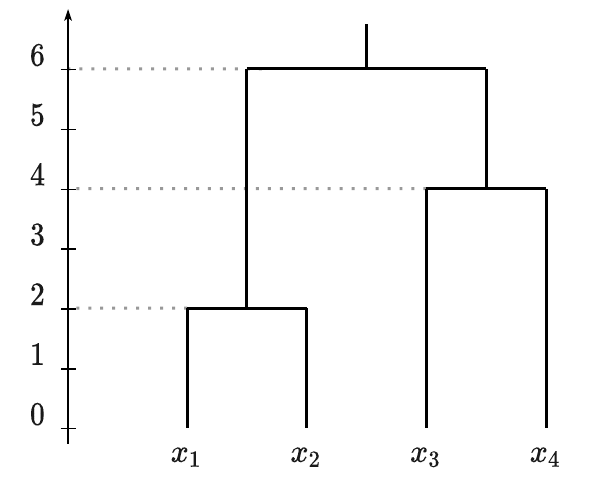}
\end{figure}

The agglomerative distances $d$ of a dendrogram are the points where two branches merge, as illustrated in Figure \ref{fig:dendro}. 

\subsection{Distance between dendrograms}

The equivalence between dendrograms and ultrametric spaces was demonstrated in~\cite{carlsson2010characterization}. Ultrametric space satisfies a stronger type of the triangular inequality. It uses the fact that, for every $r\geq0$ there is a block $\mathcal{B}$ of dendrogram $\theta(r)$ in which elements $x_i$ and $x_j$ are agglomerated, if and only if there is an ultrametric space in which the distance between those elements is smaller than or equal to $r$. Therefore there is a bijective function $\Psi: \mathcal{D}(X) \rightarrow \mathcal{U}(X)$ mapping the collection $\mathcal{D}(X)$ of all produced dendrograms of $X$ into the collection $\mathcal{U}(X)$ of all ultrametric spaces of $X$. 

Given a finite set $X$, for each produced dendrogram $\theta \in \mathcal{D}(X)$ there is a corresponding ultrametric space. The conversion between a dendrogram $(X, \theta)$ and a ultrametric space $(X, u)$, with a ultrametric distance $u$, is~\citep{carlsson2010characterization}:
\begin{equation}
    \label{eq:ultrametricdendrogram}
    u(x_i, x_j) = \min\{r \geq 0 | x_i \textrm{ and } x_j \textrm{belong to the same cluster if } \theta(r)\}.
\end{equation}

That is to say that the distance between two points in the ultrametric space is the point in the dendrogram where the branches coming from the two points meet. 

The equivalence in Equation~\ref{eq:ultrametricdendrogram} allows us to use methods from the metric space literature to analyze and better understand dendrograms for machine learning applications. For example, a method proposed by \cite{costaemploying} uses the divergence between dendrograms in order to detect concept drift on data streams. That is done by computing a approximation of the Gramov-Hausdorff divergence between the equivalents ultrametric spaces. Considering two dendrograms $(X_\alpha, \theta_\alpha)$ and $(X_\beta, \theta_\beta)$ with sorted agglomerative distances $d^{\alpha}$ and  $d^{\beta}$, the distance between the dendrograms is:
\begin{equation}
\label{eq:fausto}
    \textrm{distance}(\theta_\alpha, \theta_\beta) = \max_i |d_i^{\alpha} - d_i^{\beta}|
\end{equation}

Notice that we are using the sorted agglomerative distances as a way of aligning the clusters in the two datasets. It was proposed by \cite{costaemploying} and serves as a lower bound to the real distance between the dendrograms. For this reason we do not use an explicit assignment algorithm.

\begin{figure}[h]
\caption{Step-by-step representation of our method. (I) The set of generated images is produced by the generator. (II) A dendrogram is built using each set. (III) The metric is computed as the divergence between the dendrograms.}
\label{fig:method}
\centering
\includegraphics[width=0.5\textwidth]{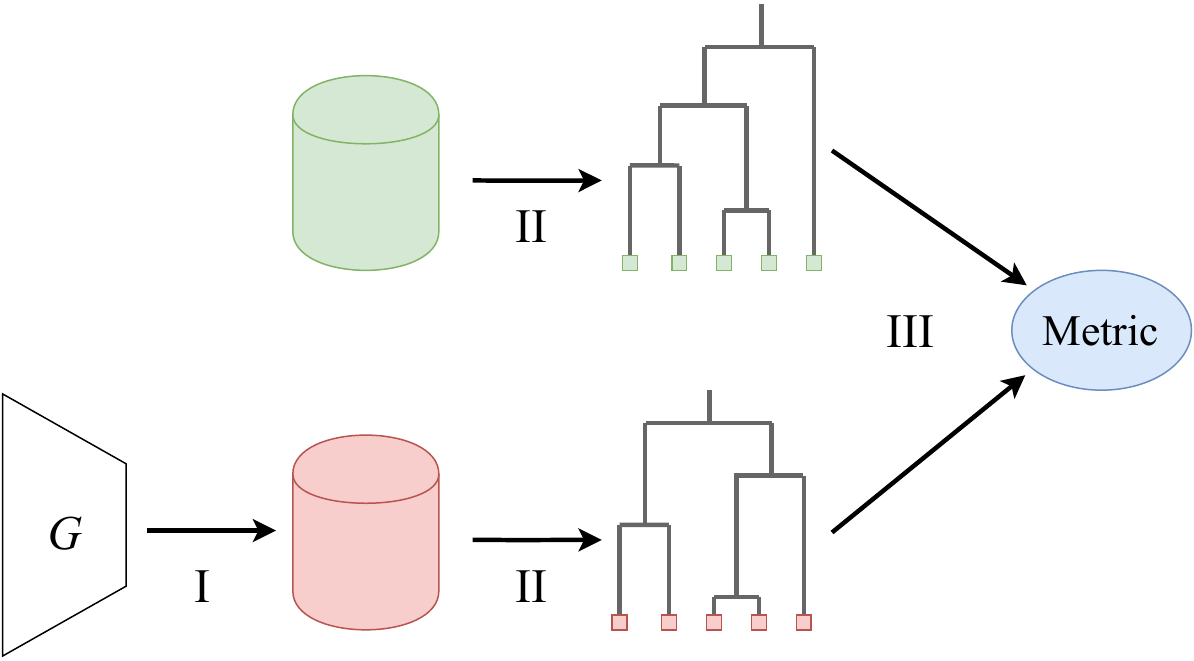}
\end{figure}

\subsection{Dendrogram Distance for Evaluation of Generative Models}

The method, illustrated in Figure \ref{fig:method}, works by computing the distance between two dendrograms, one of them constructed using samples from the real (training) data and the other from samples produced by the generator. Formally, our metric takes a set of real data $X_{data}$ and a set of generated data $X_{model}$, using them to construct their correspondent dendrograms $(X_{real}, \theta_{data})$ and $(X_{model}, \theta_{model})$ using a single-linkage clustering algorithm. The real and generated data dendrograms have agglomerative distances $d^{data}$ and $d^{model}$, respectively. Therefore, the divergence between them is computed by the following equation, which comes from Equation \ref{eq:fausto} but uses the average instead of the maximum value:
\begin{equation}
    DD(X_{data}, X_{model}) = \frac{1}{n}\sum_{i=1}^{n} |d_i^{data} - d_i^{model}|,
\end{equation}
the use of average representing a relaxation of the original formulation that allows to smooth out the results.

\subsubsection{2D Benchmarks}

Our {main} contribution is a metric to verify GANs abilities to deal with mode collapse in a fast and easy to interpret manner using 2D datasets. Such 2D datasets are widely used to visualize the mode coverage of generative models~\citep{borji2019pros}. This evaluation procedure as it is, however, relies heavily on human interpretation. Also, the data distributions are often simplistic. We propose an extension to this method, by adding more nuance to the datasets and a novel reliable quantitative measure.

\begin{figure}[h]
\caption{The two bidimensional datasets: 2D Grid and 2D Ring. They are specified by determining the mean of each mode, which are normal distributions.}
\label{fig:data}
\centering
\includegraphics[width=0.7\textwidth]{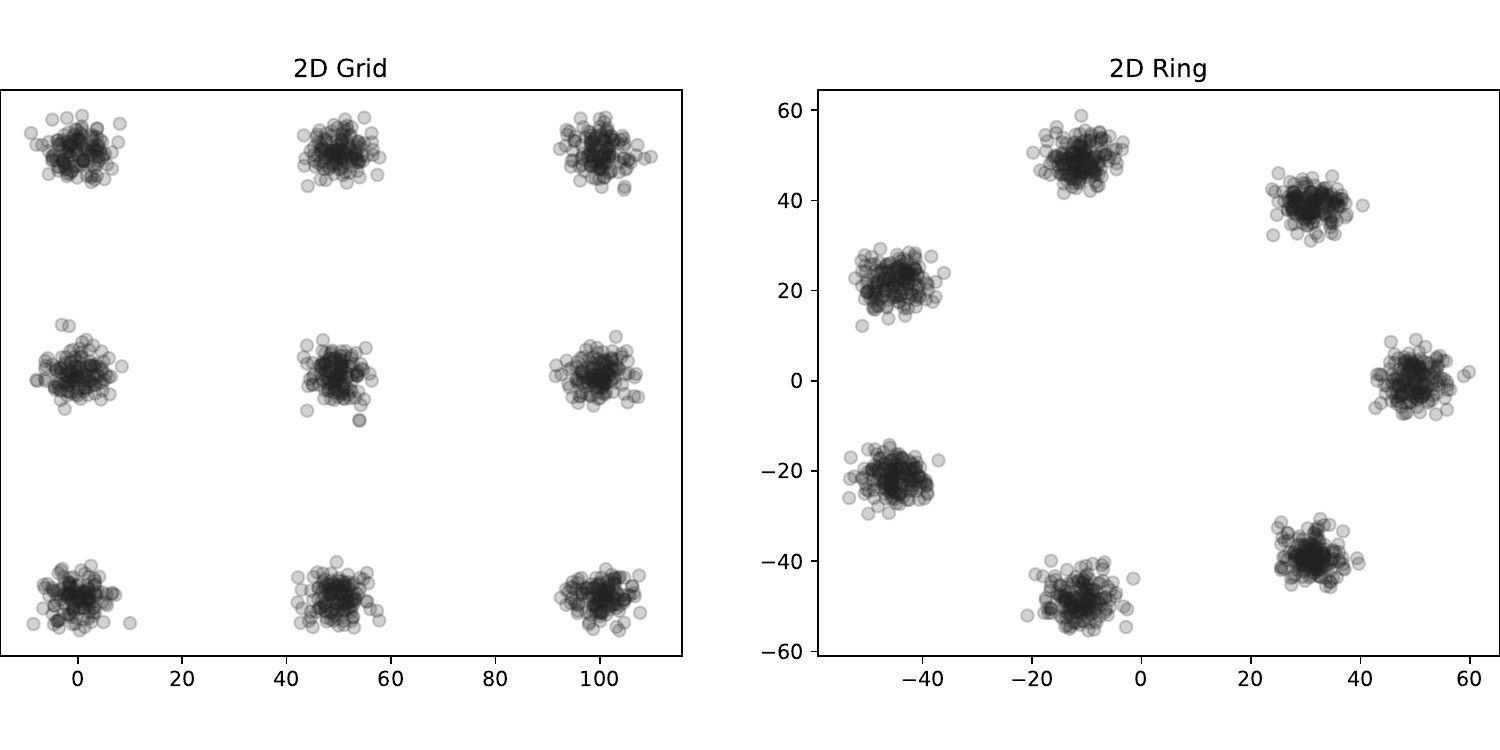}
\end{figure}

The main idea behind using synthetic 2D datasets is to easily iterate and verify the results of a model in terms of mode coverage. Two datasets are used for this purpose: the 2D Ring and 2D Grid, demonstrated in Figure~\ref{fig:data}. However, both the ring and the grid, as often employed to evaluate generative models, assume that distances between different pairs of modes are the same, which may not reflect real scenarios, such as unstructured data embeddings.

Note that the Fréchet Inception Distance (FID) between the real and generated samples makes use of mean and variance of each distribution, which we claim may not be able to properly detect the presence of mode collapse in datasets where the position of each mode is random. Our Dendrogram Distance (DD) can be used as an additional quantitative analysis of models performance on the aforementioned datasets. That is a better alternative because the DD is agnostic to the position of each mode; it only cares about the relative distance between samples. For the same reason it has a lower variance in such cases, as we further demonstrate in our experiments.

\subsubsection{Generated Image Embeddings}

Beyond its applicability on 2D datasets, the Dendrogram Distance can also be applied to other types of data, such as images. However, applying it directly to the pixel space can unfeasible due to high dimensionality. Therefore, inspired in the Inception Score (IS), which uses a pre-trained network to obtain the class distributions, we first extract a lower dimensional representation of the data points also using a pre-trained network.

In fact, both IS and FID uses this approach, which allows for the metric to invariant to certain changes in the images while preserving more information than the class distribution outputs. The architecture used as a feature extractor in this paper is a Inception-V3 pre-trained on ImageNet, in order to make our results comparable to the IS and the FID.

In summary, the Dendrogram Distance applied to unstructured data involves extracting representations for the training and generated sets, building the correspondent dendrograms with such representations and computing the divergence between the two dendrograms to get the final score. Next, we demostrate that the Dendrogram Distance produces competitive results and additional insights on image data when compared to FID and IS.

\section{Experiments and Results}

This paper includes experiments with the Dendrogram Distance (DD) in three main parts: evaluation with 2D benchmarks, with image pixels and with image embeddings.

\subsection{2D benchmark for generative models}

In this part, we employ the 2D Ring and the 2D Grid datasets. We also generated data in which the location, i.e. the mean, of each mode is perturbed, so that the relative distances between modes is not fixed.

In order to evaluate how well the metrics captured the issues in the generated distributions, we simulated generated sets with increasingly different number of modes, in order to check if the metrics reflect the changes. More concretely, the generated fake datasets that contained from 1 up to the number of modes in the true dataset. The datasets are described in Table~\ref{tab:2ddatasets}.

\begin{table}[h]
\centering
\begin{tabular}{cccc}
\textbf{Dataset} & \textbf{Target \# of Modes} & \textbf{Radius} & \textbf{Grid length} \\ \hline
2D Grid          & 9                        & --              & 100                 \\
2D Ring          & 7                        & 50              & --                  \\
\end{tabular}
\caption{Description of the 2D datasets}
\label{tab:2ddatasets}
\end{table}

Our first experiment consisted of adding different levels of noise to the position of each mode and checking if the metrics were still able to correctly quantify the number of modes in the datasets. In practice, for each mode we generated a displacement value sampled from $\textrm{Unif}(-\alpha L, \alpha L)$, where $\alpha$ is the noise scale that we varied and $L$ is the length of the grid and the diameter of the ring.

\begin{figure}[h]
\caption{Results of FID and DD on 2D Grid. The first row shows the dataset; the second and third row shows, respectively, the FID and DD results when different number of modes of the dataset are generated. For the perturbed datasets we also show in red the curve relative to the noise-free scenario.}
\label{fig:grid-noise}
\centering
\includegraphics[width=\textwidth]{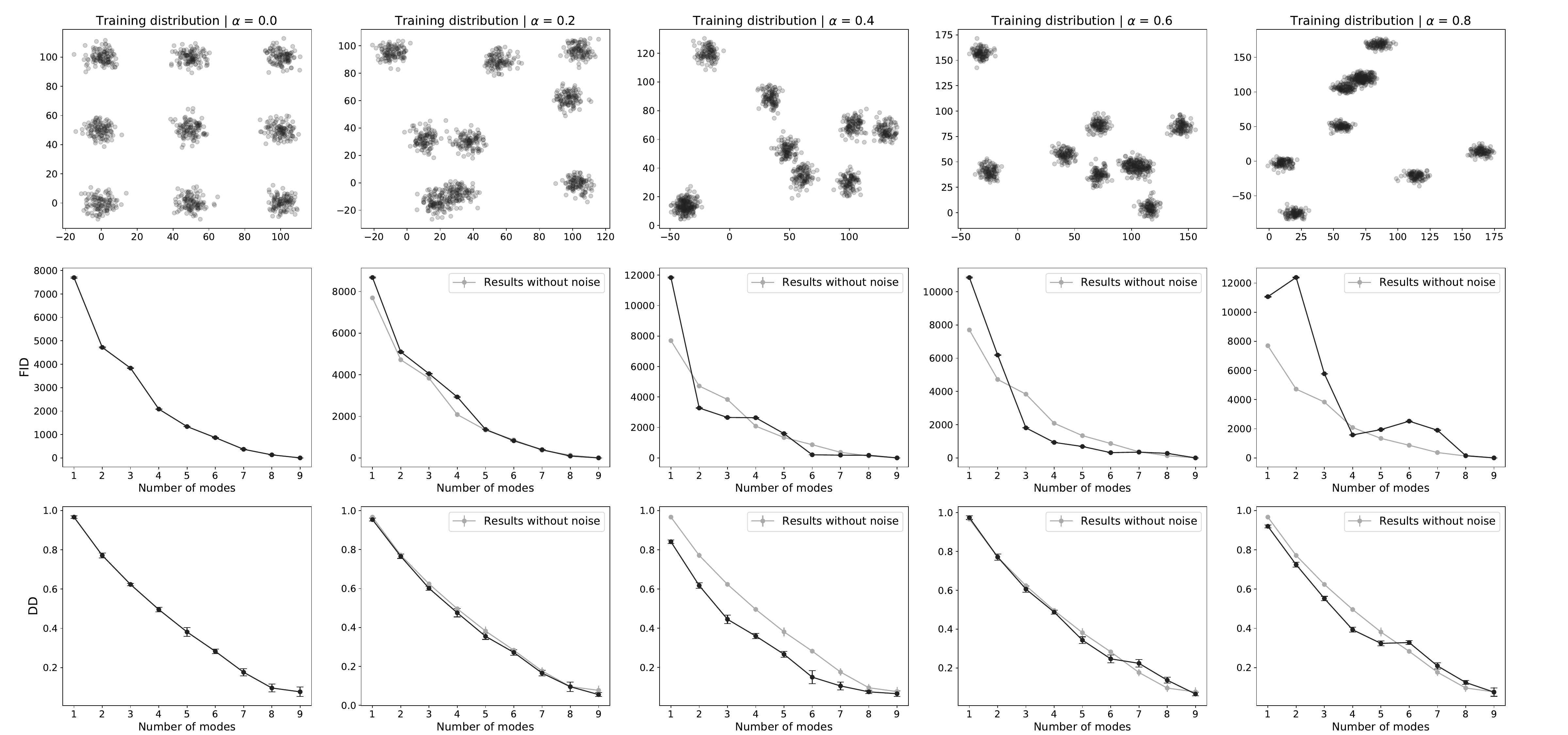}
\end{figure}

\begin{figure}[h]
\caption{Results of FID and DD on 2D Ring. The first row shows the dataset; the second and third row shows, respectively, the FID and DD results when different number of modes of the dataset are generated. For the perturbed datasets we also show in red the curve relative to the noise-free scenario}
\label{fig:ring-noise}
\centering
\includegraphics[width=\textwidth]{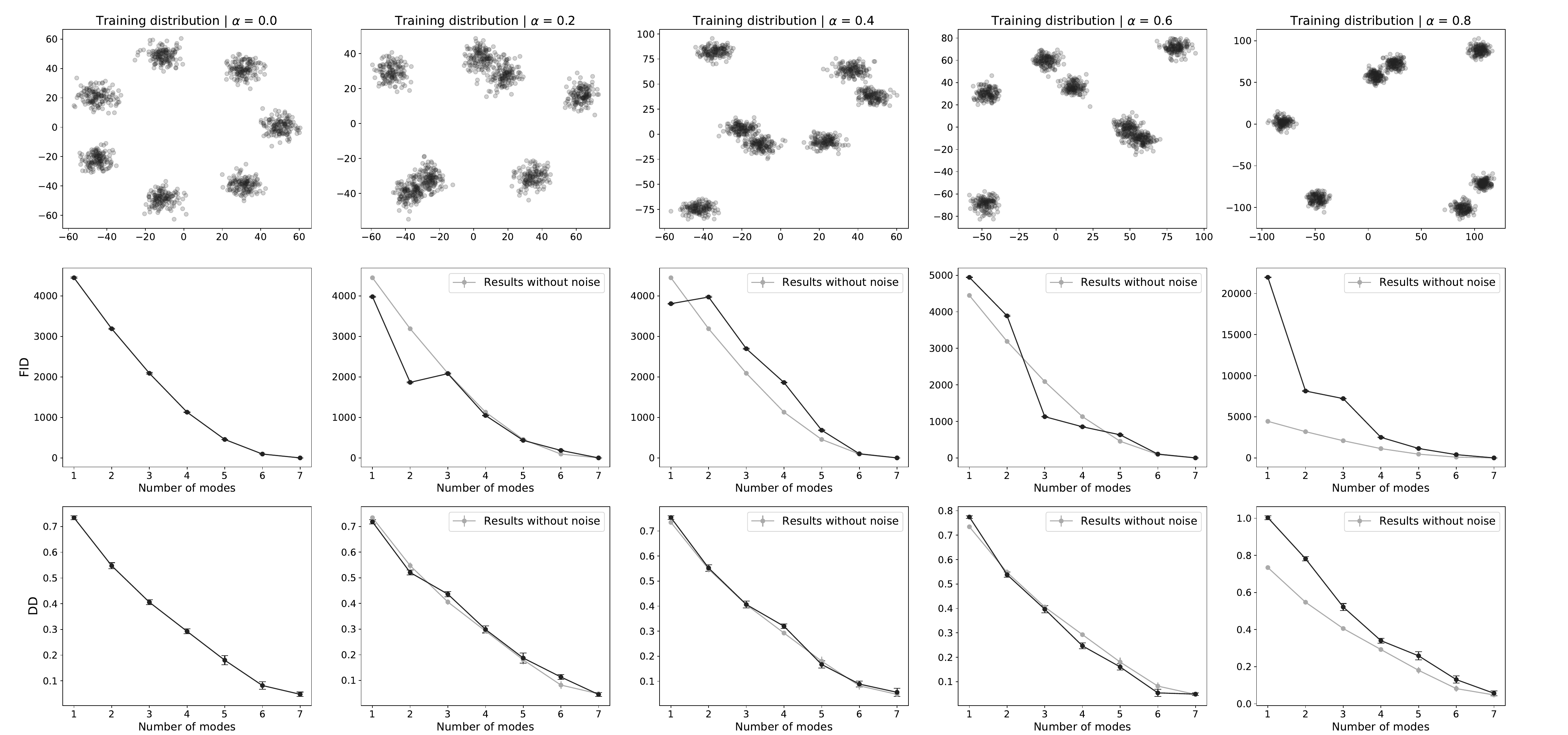}
\end{figure}

In Figure~\ref{fig:grid-noise} we show a simulation with 2D Grid in which we add modes iteratively and compute both FID and DD. In a noise-free scenario, both DD and FID captures well the number of modes, but, as the mean of each mode is perturbated, the FID suffers from instability, while the DD is more stable and smooth, therefore demonstrating to be more robust. Similarly, in Figure~\ref{fig:ring-noise} the simulation with 2D Ring is shown, with similar outcomes: DD grows more smoothly as the number of modes increase.

We also performed experiments to check if the Dendrogram Distance has a lower variance when compared to the FID. For that we continue to use the strategy with different noise factors, but we executed the experiment ten times for each factor and mode. By doing that we are able to verify if the metrics assign different scores for the same number of modes if their positions change. 

\begin{figure}[h]
\caption{2D Grid stability analysis for different perturbations of the modes.}
\label{fig:grid-std}
\centering
\includegraphics[width=\textwidth]{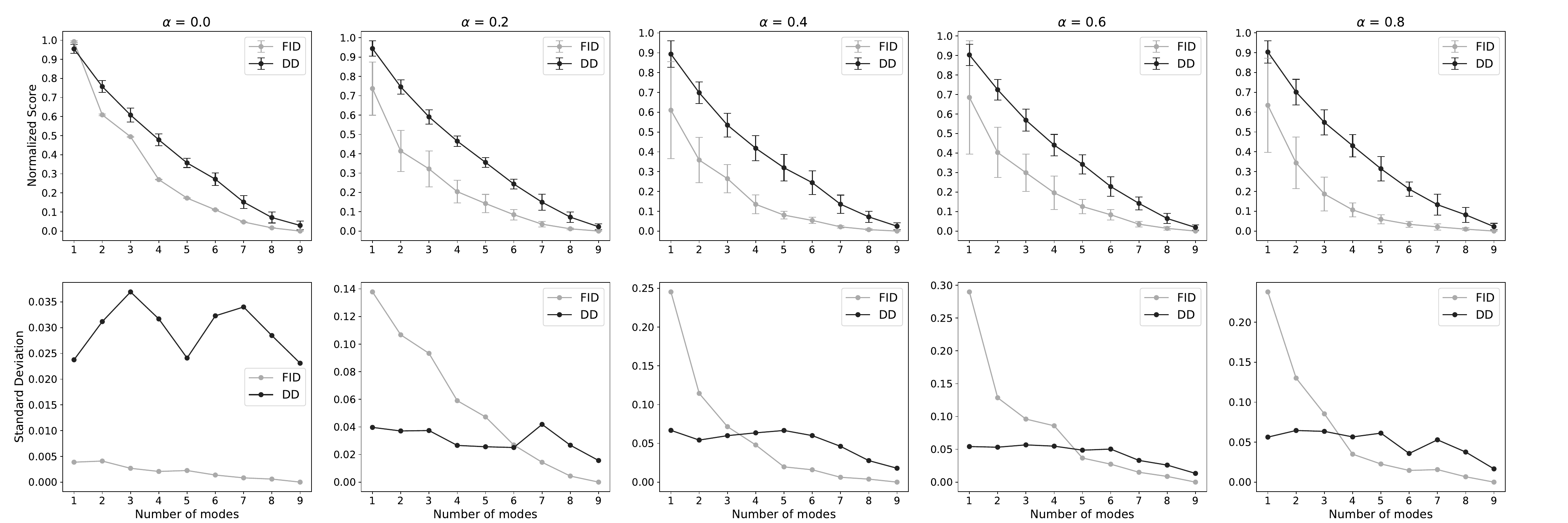}
\end{figure}

\begin{figure}[h]
\caption{2D Ring stability analysis for different perturbations of the modes}
\label{fig:ring-std}
\centering
\includegraphics[width=\textwidth]{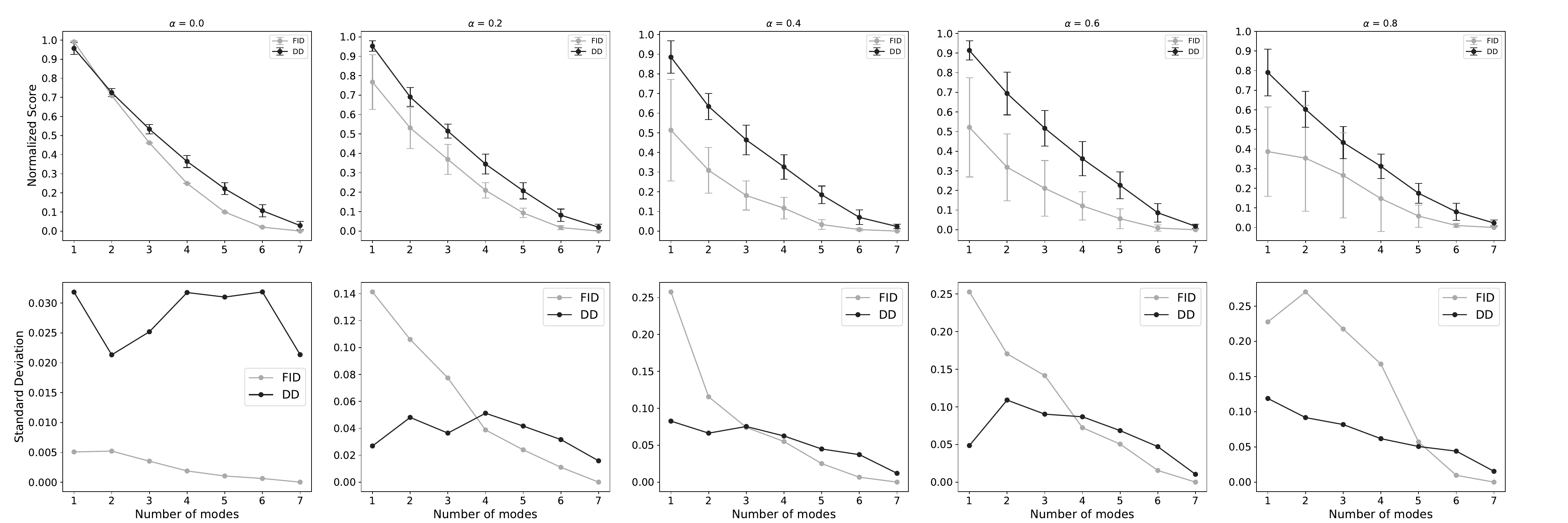}
\end{figure}

As shown in Figure~\ref{fig:grid-std} for the 2D Grid and Figure~\ref{fig:ring-std} for the 2D Ring, in the noise-free scenario the DD still captures better the presence of new modes, with both DD and FID showing small standard deviations. As the noise increases, the FID becomes unstable with respect to the value of the metric in particular when more modes are present, while the DD metric is less affected by the number of modes.

Those results indicate the advantages of the Dendrogram Distance for capturing mode collapse in 2D datasets. It presents lower standard deviation as the number of modes dropped get higher and the randomness in the position of the modes increase, both cases in which a good evaluation becomes even more important. Conversely, that is not the case for the Frechet Inception Distance. The results, overall also demonstrate that the Dendrogram Distance is more sensitive with respect to the number of modes than the FID. It also shows that, if the representation of the data is such that the modes are well defined in the features space, then the DD suits very well the evaluation task.

\subsection{Image Pixels}

This part investigates the use of DD directly in image pixels. We used MNIST dataset due to its relatively lower dimensionality and discriminative properties of its pixel space.

\subsubsection{Mode collapse detection}

The intuition for using Dendrogram Distance is to better capture the underlying distribution of the data, therefore making it suitable to detect mode collapse. In order to assess this ability on image datasets, we first create a proof-of-concept by sampling from a real dataset to simulate training and generated data points.

More precisely, we start by sampling the training set which contains all the modes. Then, using the same source dataset, we sample an increasing number of modes, by taking into account the labels of the images. For this reason we use a classification dataset. We drop one mode at a time, going from ten to one modes in each simulated generated set. This procedure is performed ten times to better understand the method's variance.

\begin{figure}[h]
\caption{Behavior of the Dendrogram Distance as the number of modes in the data changes. It is possible to notice that as the number of modes gets lower the scores became worse.}
\label{fig:modes-mnist}
\centering
\includegraphics[width=0.7\textwidth]{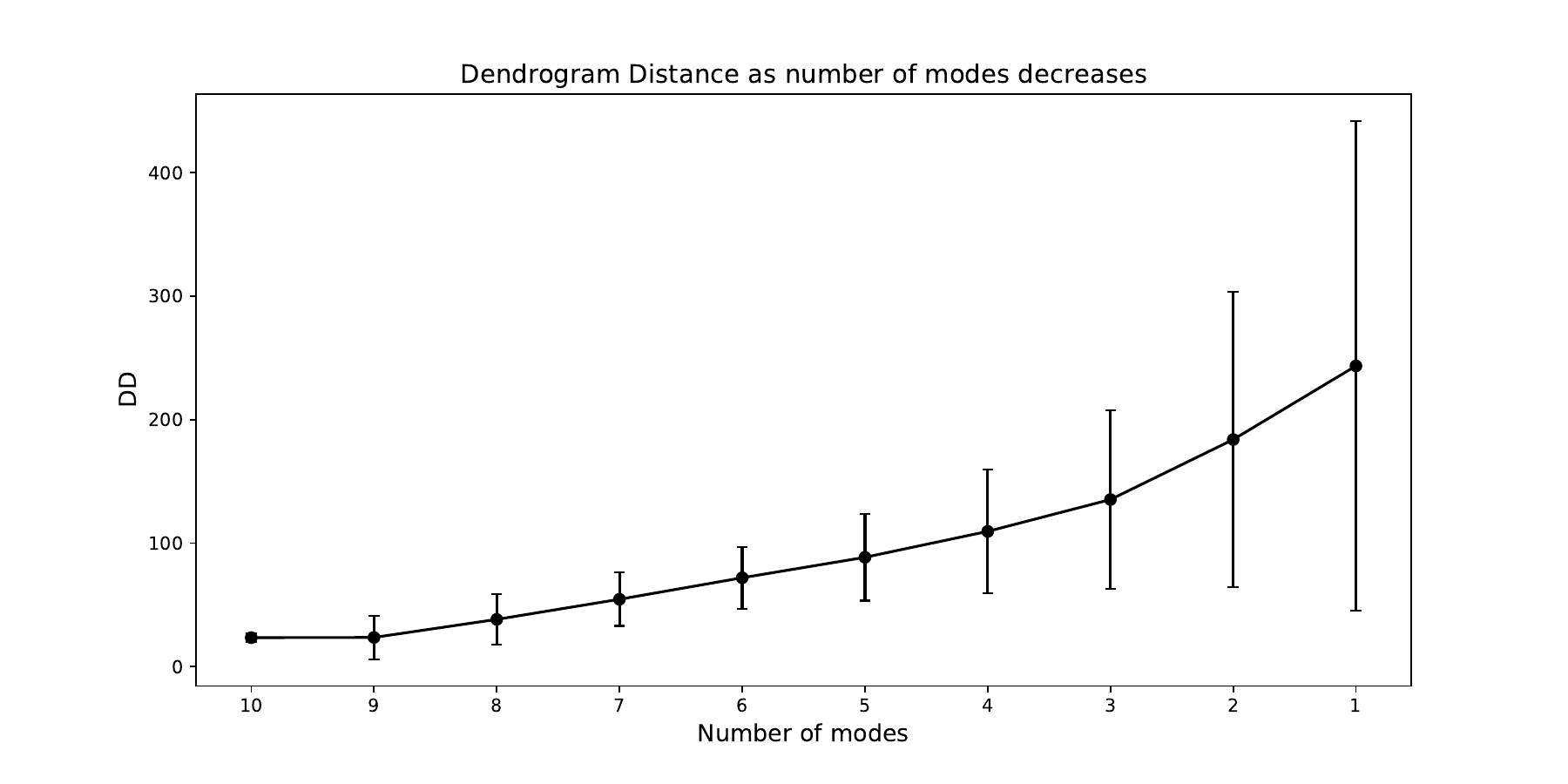}
\end{figure}

The results, presented in Figure \ref{fig:modes-mnist}, demonstrate that the Dendrogram Distance is able to correctly detected mode collapse using MNIST's pixel space. The presented curve shows that the metric is able to distinguish between different levels of mode collapse properly. It is important to point out that the metric does have a high standard deviation, which is a consequence of working directly with pixels.

\subsubsection{Metric as convergence proxy}

We also examine if the Dendrogram Distance is able to detect the convergence of a generative model during training. That is relevant given that GANs are optimizing two objectives at the same time and to analyze convergence is not as easy as in other learning tasks. 

\begin{figure}[h]
\caption{Dendrogram Distance changes during training and a corresponding sample at each iteration.}
\label{fig:wgan-mnist}
\centering
\includegraphics[width=0.8\textwidth]{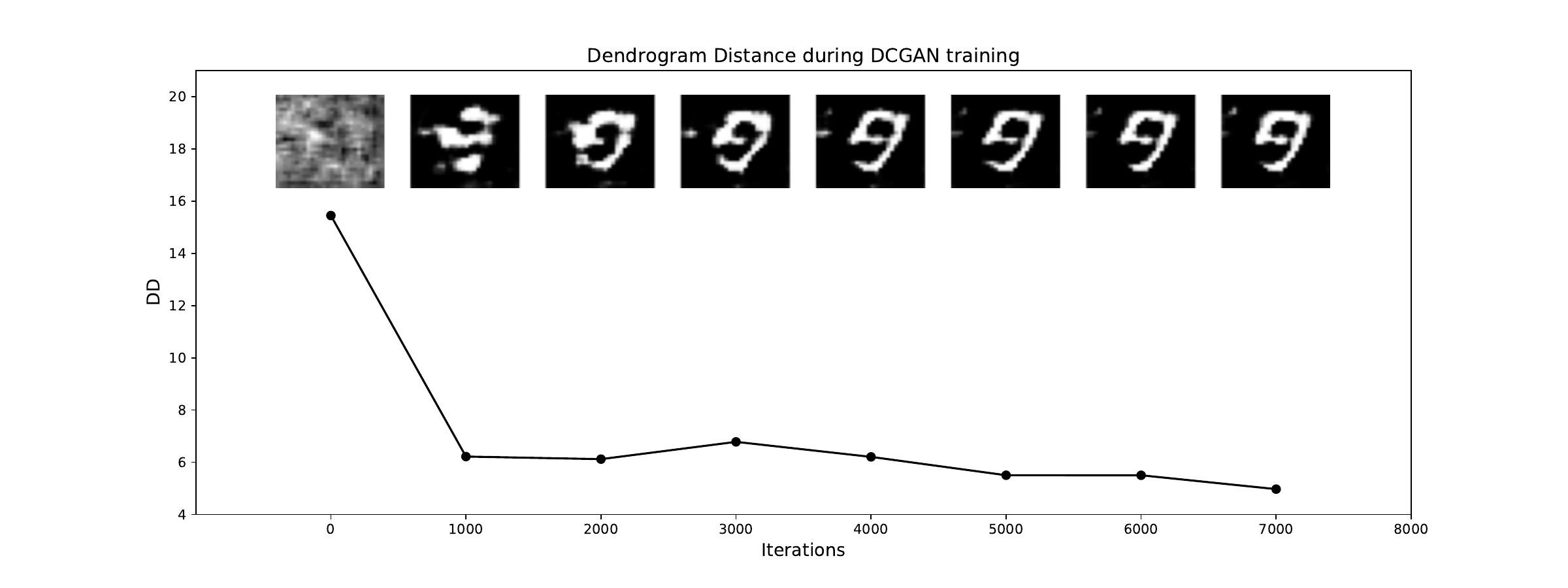}
\end{figure}

To this end, we trained a WGAN \citep{arjovsky2017b} on MNIST for 7000 iterations using RMSprop with learning rate $10^{-5}$. During training we kept track of its generated samples, computing the DD during training to see how it relates with overall sample quality. Our results are shown in Figure \ref{fig:wgan-mnist}.

These results demonstrate that the Dendrogram Distance is able to indicate the generator convergence. The metric correctly follows the quality of the generated samples. Some fluctuations in the metric are a result of the instability during the training process of GANs, due to the two adversarial objectives.

\subsection{Generated Image Embeddings}

Finally, we studied the use of image embeddings, obtained from pre-trained CNNs, to evaluate generated data, making it viable to assess datasets with higher dimensionality. In particular, we focus on detection of mode collapse, as it is the main focus of the proposed metric.

We adopt the same sampling strategy used for MNIST on three image classification datasets: CIFAR-10, STL-10 and CelebA. On the first two each class was said to be a different mode of the data distributing, therefore on both datasets ten modes are considered. For the simulated generated sets we drop the number of modes from ten to one. On CelebA there is no explicit class label for the images, thus we constructed four sets using the attributes: dark-haired men, blonde men, dark-haired women and blonde women. We drop one mode at a time, going from four to one in the order just presented.

We compare the Dendrogram Distance (higher value is better) with the Fréchet Inception Distace (higher value is better) and the Inception Score (lower value is better). As mentioned before we used a Inception-V3 as a feature extractor, resizing all images to $299 \times 299$ before passing them through the network. The Dendrogram Distance is using the output of the final convolutional layer of the network to represent the data points.

\begin{figure}[h]
\caption{Results on CIFAR-10}
\label{fig:modes-cifar}
\centering
\includegraphics[width=1\linewidth]{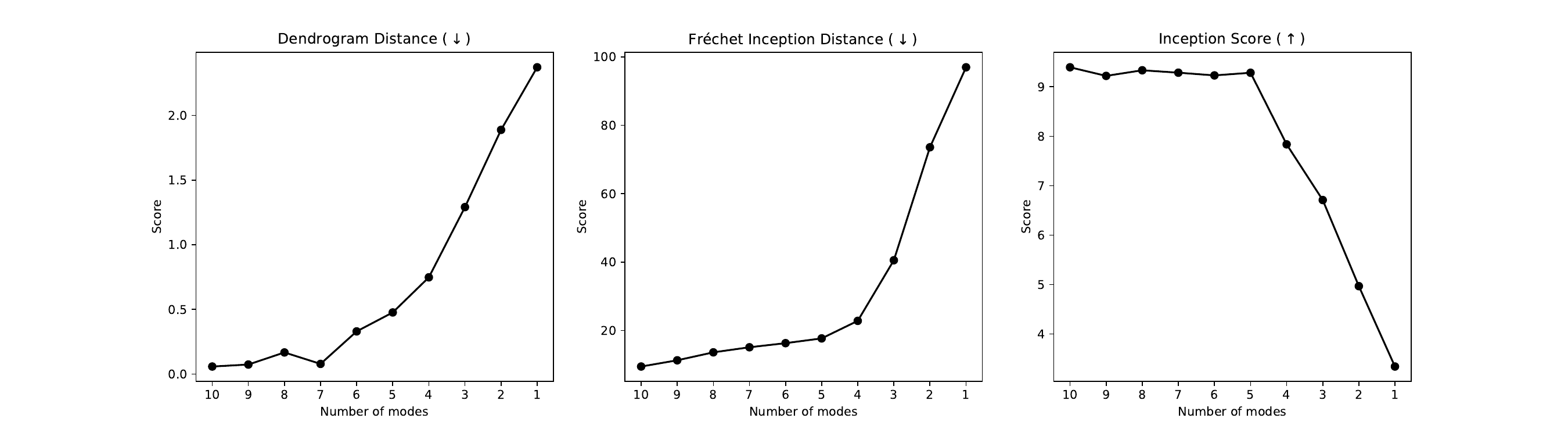}
\end{figure}

The results for CIFAR-10 can be seen in Figure \ref{fig:modes-cifar}. The DD was capable of capturing the mode collapse, outputting high values for 1, 2 and 3 modes, and increasingly lower values when more modes were present. The FID was more sensitive on higher number of modes, while the IS failed to detect new modes up until 7 modes appeared in the generated dataset.

\begin{figure}[h]
\caption{Results on STL-10}
\label{fig:modes-stl}
\centering
\includegraphics[width=\textwidth]{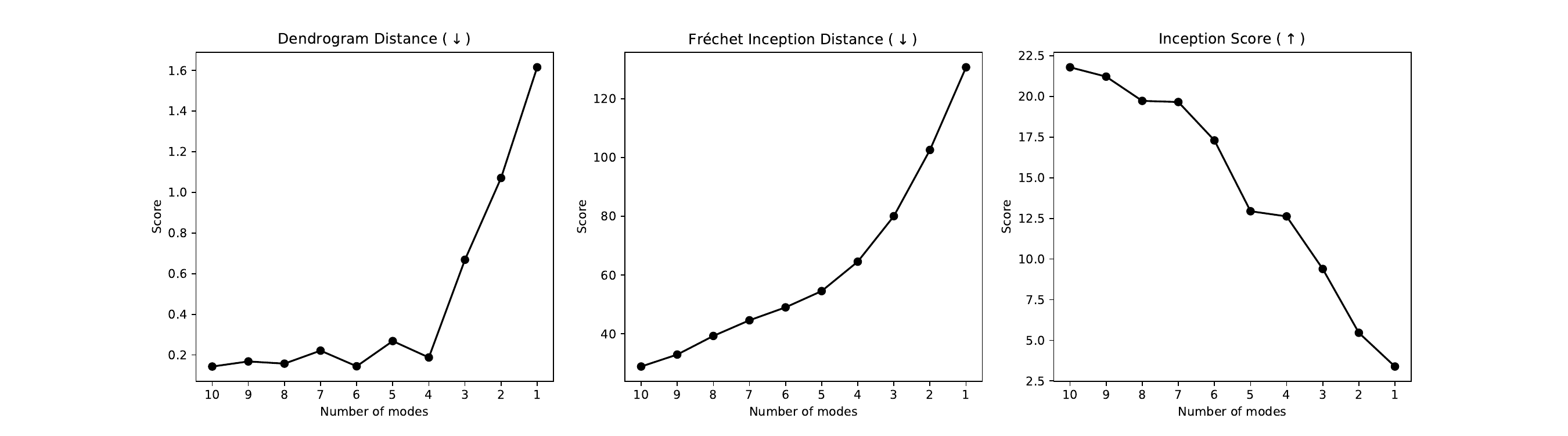}
\end{figure}

When analyzing the results on STL-10 in Figure \ref{fig:modes-stl} is possible to see that the DD had more difficult to distinguish between a higher number of modes when compared to the FID. On this dataset the IS presented a better performance, being able to distinguish between the different number of modes.

\begin{figure}[h]
\caption{Results on CelebA}
\label{fig:modes-celeba}
\centering
\includegraphics[width=\textwidth]{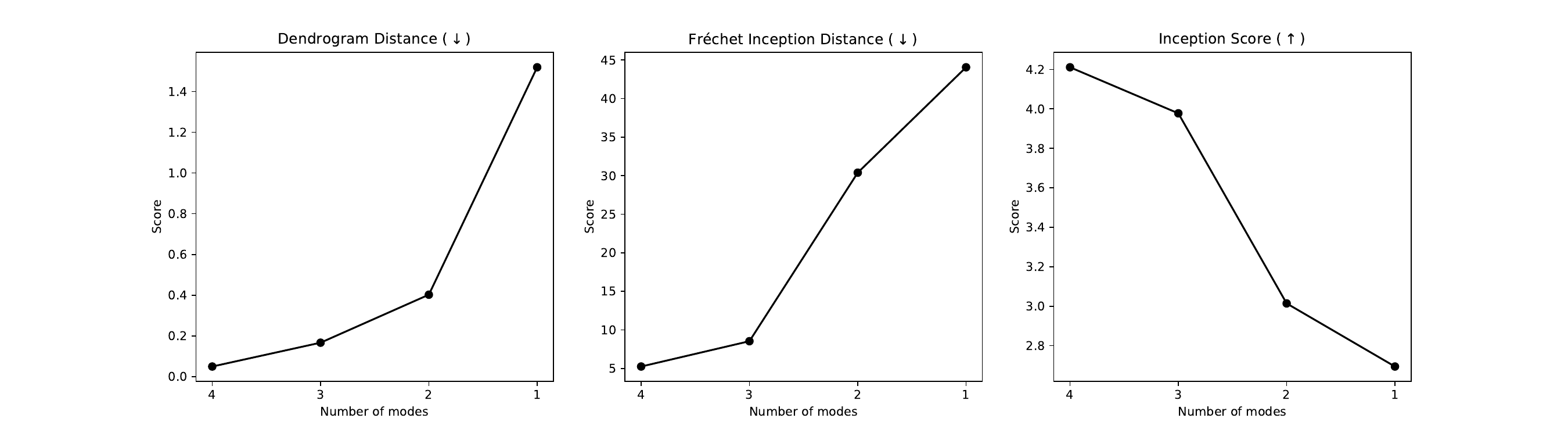}
\end{figure}

All of the metrics were able to perform well on CelebA, as can be seen in Figure \ref{fig:modes-celeba}. The results do not differ by much in terms of mode collapse detection. We hypothesize that it happens because the different between the modes are presented very explicitly in the image, given that relevant features for, like hair color and length, represent a large portion of the images.

The overall result of the Dendrogram Distance demonstrates that despite not being developed with image evaluation in mind, the metric is capable of dealing with such type of data.  It is worth mentioning that the DD depends on the formation of clusters, so we believe that, with better feature spaces, we can achieve even better results.

\section{Conclusions}

The Dendrogram Distance has demonstrated to be a stable approach for quantifying mode collapse on synthetic bi-dimensional datatets. The use of dendrograms makes the metric agnostic to the position of the modes, allowing it to deal with more complex distributions where the distance between groups of data points may vary. By providing a more reliable quantitative measure for the 2D Ring and 2D Grid the Dendrogram Distance can be established as a useful benchmark when developing novel generative models. Also, the Dendrogram Distance is capable of dealing with image data, both in pixel and feature spaces. The quality of the results does confirm the capabilities of the method. The DD allows for easy verification of the results while offering interpretability and theoretical foundations linking clusters and ultrametric spaces. 

Considering the performance on the bi-dimensional regime, it indicates that modification in the metric pipeline might lead to better results when dealing with images. For instance, improving the feature extraction to make it work better for the clustering task should yield better results. The results also suggest that is worth investigating the use of the Dendrogram Distance as an additional training objective, instead of being used just as an evaluation metric. This and other improvements, such as using DD as a general metric for comparison between datasets, are beyond the scope of this paper, but can be a matter of investigation in future work.

\section*{Acknowledgements}
This work is supported by FAPESP grants \#2018/22482-0 and \#2019/07316-0; CNPq fellowship \#304266/2020-5.
\bibliography{refs}   


\end{document}